\documentclass[default,iicol]{sn-jnl}% Default with double column layout

\usepackage{booktabs}       % professional-quality tables
\usepackage{amsfonts}       % blackboard math symbols
\usepackage{nicefrac}       % compact symbols for 1/2, etc.
\usepackage{microtype}      % microtypography
\usepackage{lipsum}
\usepackage{fancyhdr}       % header
\usepackage{graphicx}       % graphics
\graphicspath{{media/}}     % organize your images and other figures under media/ folder
\usepackage{amsmath}
\usepackage{subcaption}
\usepackage{wrapfig}
\usepackage{float}
\usepackage{xspace}
\usepackage{enumitem}
\usepackage{xcolor}

\pdfoutput=1

\usepackage[ruled, lined, linesnumbered, commentsnumbered, longend]{algorithm2e}

\newcommand{\ie}                {\emph{i.e.},\xspace}

\newcommand{\etal}              {\emph{et~al.}\xspace}

\definecolor{darkgreen}{rgb}{0,0.5,0}
\definecolor{purple}{rgb}{0.75,0,0.75}
\definecolor{brown}{rgb}{0.65,0.16,0.16}
\definecolor{darkslateblue}{rgb}{0.28, 0.24, 0.55}
\definecolor{orange}{rgb}{1.0, 0.647, 0}

\newcommand{\abs}[1]{\lvert #1 \rvert}

\newcommand{\us}{\small\textsc{S-TrAdaBoost.R2}\normalsize}

\newcommand{\tradaR}{\small\textsc{TrAdaBoost.R2}\normalsize}
\newcommand{\adaR}{\small\textsc{AdaBoost.R2}\normalsize}

\newcommand{\trada}{\small\textsc{TrAdaBoost}\normalsize}
\newcommand{\ttradaR}{\small\textsc{TTR2}\normalsize}

\newcommand{\kmmx}{\small\textsc{KMM.TL}\normalsize}
\newcommand{\kliepx}{\small\textsc{KLIEP.TL}\normalsize}
\newcommand{\iwkrrx}{\small\textsc{IW-KRR.TL}\normalsize}

\jyear{2021}%

\theoremstyle{thmstyleone}%
\theoremstyle{thmstyletwo}%

\theoremstyle{thmstylethree}%

\begin{document}

% \title[\us{}]{Boosting For Regression Transfer via Importance Sampling}
\title[\us{}]{ISTRBoost: Importance Sampling Transfer Regression using Boosting}


\author[1]{\fnm{Shrey} \sur{Gupta}}\email{shrey.gupta@emory.edu}

\author[2]{\fnm{Jianzhao} \sur{Bi}}\email{jbi6@uw.edu}

\author[3]{\fnm{Yang} \sur{Liu}}\email{yang.liu@emory.edu}

\author*[1]{\fnm{Avani} \sur{Wildani}}\email{avani@mathcs.emory.edu}

\affil*[1]{\orgdiv{Department of Computer Science}, \orgname{Emory University}, \orgaddress{\city{Atlanta}, \state{GA}, \country{US}}}

\affil[2]{\orgdiv{Department of Environmental and Occupational Health Sciences}, \orgname{University of Washington}, \orgaddress{\city{Seattle}, \state{WA}, \country{US}}}

\affil[3]{\orgdiv{Gangarosa Department of Environmental Health}, \orgname{Emory University}, \orgaddress{\city{Atlanta}, \state{GA}, \country{US}}}


\abstract{Current Instance Transfer Learning (ITL) methodologies use domain adaptation and sub-space transformation to achieve successful transfer learning. However, these methodologies, in their processes, sometimes overfit on the target dataset or suffer from negative transfer if the test dataset has a high variance. Boosting methodologies have been shown to reduce the risk of overfitting by iteratively re-weighing instances with high-residual. However, this balance is usually achieved with parameter optimization, as well as reducing the skewness in weights produced due to the size of the source dataset. While the former can be achieved, the latter is more challenging and can lead to negative transfer. We introduce a simpler and more robust fix to this problem by building upon the popular boosting ITL regression methodology, two-stage TrAdaBoost.R2. Our methodology,~\us{}, is a boosting and random-forest based ensemble methodology that utilizes importance sampling to reduce the skewness due to the source dataset. We show that~\us{}~performs better than competitive transfer learning methodologies $63\%$ of the time. It also displays consistency in its performance over diverse datasets with varying complexities, as opposed to the sporadic results observed for other transfer learning methodologies.}



\maketitle

\section{Introduction}
%talk about consistency
%Motivation should be more strong
%Give examples why the problem matters
%Start breakdown from the bigger problem to the technicalities

%Why transfer learning matters.
%What fields can use transfer learning.
While semi-supervised learning and unsupervised learning methodologies work well for partially labelled or unlabelled datasets~\cite{oliver2018realistic, bengio2012deep}, they fall short for instances where sample size is small~\cite{wang2020generalizing, weiss2016survey, pan2009survey, wei2018uncluttered, finn2017model}. Instance Transfer Learning (ITL)~\cite{weiss2016survey, pan2009survey, wei2018uncluttered, day2017survey, guan2015local, obst2021transfer, du2016hypothesis}, a sub-class of data-based transfer learning approaches~\cite{zhuang2020comprehensive}, is designed for limited and labelled samples, shared feature-space,  and independent and identically distributed (i.i.d) data-distributions~\cite{pan2010transfer, taylor2009transfer}, making it ideal for real-world datasets~\cite{cheplygina2019not, karpatne2018machine, chen2016spatially, qi30deep, mei2014adaboost, lv2019air}. It stands apart from its counterparts, such as Feature transfer learning and Parameter transfer learning, as it allows data adjustment and transformation of domain instances, making it ideal for dissimilarly distributed source and target domains. Moreover, ITL methodologies are as statistically interpretable~\cite{chen2016xgboost} as they are powerful~\cite{wei2018uncluttered, blanchard2017domain}, which increases their usability for domain experts~\cite{wang2019instance} who avoid complex, black-box methodologies~\cite{bashar2020regularising, han2018new, jing2019task}. Therefore, these methodologies have an advantage of being less complex but equally reliable when compared to deep transfer methodologies. Another reason for leaning towards ITL methodologies is because it is easier to transfer the source domain by applying adaptation methodologies~\cite{huang2006correcting, sugiyama2008direct} as well as using techniques involving reduction of distribution difference between the source and the target domain~\cite{cortes2014domain, garcke2014importance, sugiyama2008direct}. The accuracy of prediction does not just depend on the transfer learning methodology, but also involves the nature of distribution. Real-world datasets suffer from collecting data that is complete, high-resolution, and evenly sampled. This is due to the dependence on cost of equipment which can result in hardware limitations. This leads to the resulting dataset varying in resolution as well as the quality~\cite{lv2019air}. Hence, a robust transfer learning methodology should perform consistently well for data distributions with varying complexities.   

Among the ITL methodologies, we employ ensemble methodology, especially the boosting methodology~\cite{chen2016xgboost} as it aggregates the results from multiple learners. Similarly, the transfer boosting methodology~\tradaR{}~\cite{pardoe2010boosting} is regularized and uses domain adaptation for iteratively re-weighing the source instances with respect to the target dataset for knowledge transfer~\cite{tang2020improving}. The underlying architecture is AdaBoost~\cite{freund1999short}, which focuses on misclassified traning instances, leading to a contextual learning. However, boosting methodologies suffer from negative transfer~\cite{rosenstein2005transfer} when source dataset size is large compared to the target dataset, leading to a skewed final model.
To address the problem of negative transfer, we introduce~\us{}, a successor to two-stage TrAdaBoost.R2 (\ttradaR{}) that uses importance sampling~\cite{pan2009survey, katharopoulos2018not, zhao2015stochastic} to improve alignment of source instances with the target values, and also mitigates the skewness generated due to the large sample size of source datasets. We test \us{}~across a range of standard regression datasets with limited target instances and varying complexities, and find that it outperforms other ITL methodologies $63\%$ of the times and the baseline~\ttradaR{} more than $75\%$ of the times. Notably, it has a consistent performance (RMSE and R-squared score) for both the regular comparative study and the Ablation study (Fig.~\ref{fig:uci-rmse-r2} and Table~\ref{tab:abaltion-stdreg}), as opposed to fluctuating results observed for other methodologies. 
 
 The primary contributions of this paper are:
 \begin{enumerate}
    \item We introduce~\us, a complexity-tolerant, domain-agnostic boosting-based transfer learning algorithm that uses importance sampling and an unconstrained weight update strategy to outperform its predecessor~\ttradaR{} and other competitive ITL methodologies. 
    \item We discuss the complexity measures, \ie, metrics to quantify complexity of distribution. They categorize the distribution based on correlation, linearity, and smoothness, to provide a numerical estimate of its simplicity.
    \item We demonstrate that~\us{} outperforms competitive ITL methodologies when measured in terms of accuracy and loss, for high complexity datasets. We also provide the ablation analysis for Importance Sampling, which demonstrates the modularity and commutability of the technique.
\end{enumerate} 

% In the remainder of this paper, we provide more insight into the problem domain, describe \us{}~in detail, discuss what it means for a dataset to be complex, and, finally, demonstrate successful transfer in complex datasets through the use of~\us{}. The Background section sheds more light over negative transfer and evolution of boosting methodologies to cater to transfer learning. We describe the problem statement and introduce the~\us{} and k-Center Sampling in the Methodology section. We discuss the results comparing ITL methodologies for numerous datasets in the Evaluation section and conclude our study with performance takeaways and suggestions for the future work.
 
\section{Background}
\begin{figure}[ht]
 \centering 
 \includegraphics[width=\columnwidth]{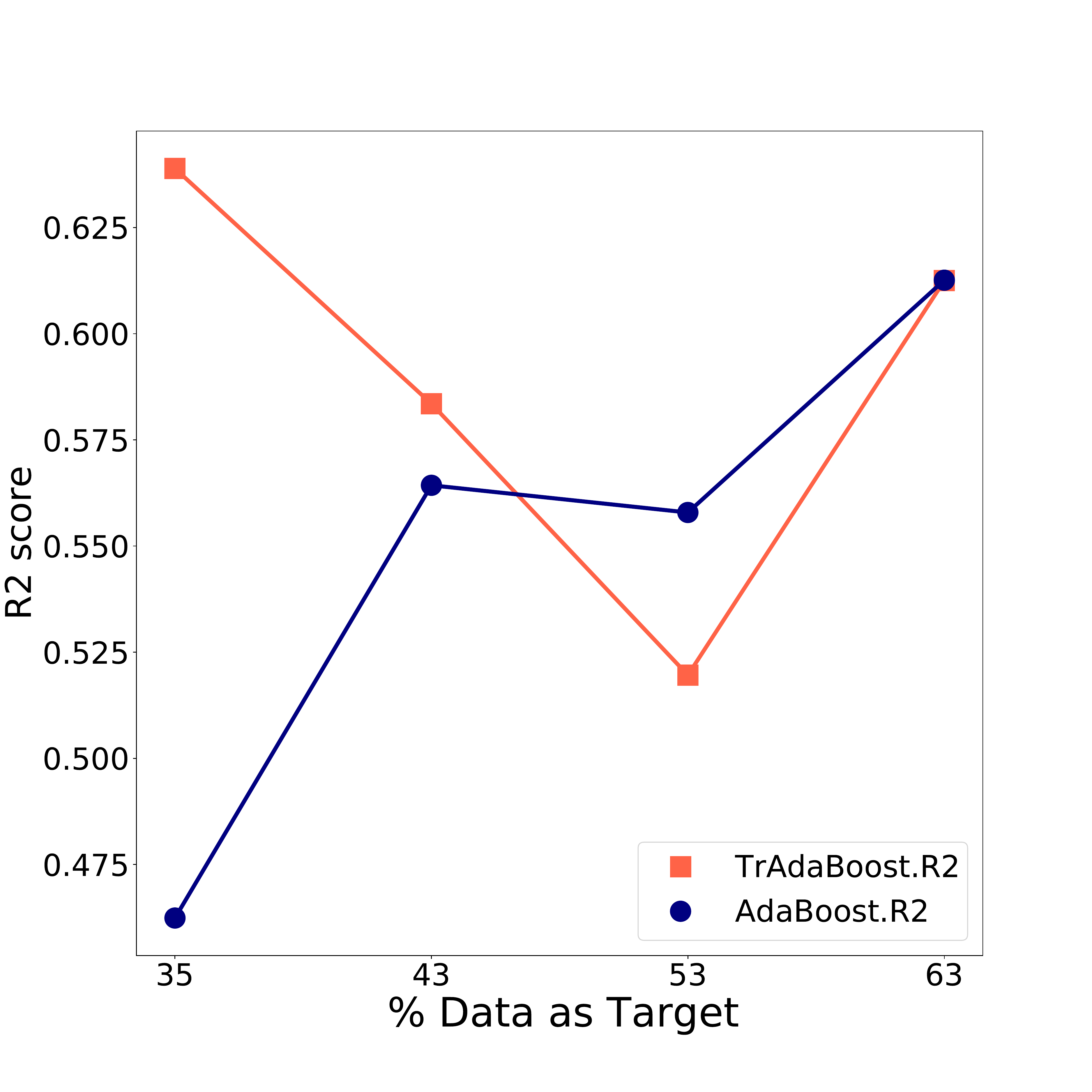}
%  {paper_negative_transfer.eps}
%  {paper_negative_transfer.png}
 \caption{Negative transfer in~\ttradaR{} is induced as a result of increasing the target sample size from 35\% to 63\% of the total training data. The baseline algorithm is~\adaR{}. For a larger target sample size, the baseline performs better than~\ttradaR{}}
 \label{fig:neg_transfer}
\end{figure}

Previous work on transfer learning~\cite{day2017survey, weiss2016survey} provides methodologies for measuring the shared information content between multiple domains in transfer learning~\cite{mahmud2007transfer,mihalkova2007mapping,swarup2006cross}. These models attempt to find common structural representations of source instances to gauge the quantity as well as quality for the transfer. However, for highly dissimilar source and target domain instances, a reduction of prediction accuracy for transfer learning algorithms when compared to non-transfer learning algorithms \ie \emph{negative transfer} is commonplace~\cite{rosenstein2005transfer}. Fig~\ref{fig:neg_transfer} shows negative transfer when~\ttradaR{} and~\adaR{} are fitted over the Concrete dataset from UCI machine learning repository~\cite{asuncion2007uci}. We observe a decline in~\ttradaR{}'s performance as the target sample size increases. This shows a trade-off in the performance of transfer learning algorithms to the sample size of the target distribution. Hence, transfer learning algorithms perform better when the sample size of a target dataset is small.

The concept of translating knowledge and model across domains has been much researched upon and hence, transfer learning, similar to machine learning, is observed for both classical transfer learning~\cite{cao2010adaptive,cortes2014domain,garcke2014importance,pardoe2010boosting} and deep transfer learning methodologies~\cite{bashar2020regularising,han2018new,jing2019task,ma2019improving,tan2018survey,yao2019learning,zhang2019interactive,zuo2016fuzzy}. While deep networks can often improve transfer accuracy, they sacrifice model interpretability, generalizability, adaptability, and flexibility for more diverse tasks~\cite{camilleri2017analysing,ramakrishnan2016towards}. Whereas, ITL algorithms, unlike deep transfer models, do not suffer from obscurity in showing intermediary steps and learnt concepts in order to have a greater transparency. Even for unrelated source and target domains, the source instances adapt to the target instances by either re-weighting~\cite{cao2010adaptive,garcke2014importance} or transforming to the target space~\cite{cortes2014domain}, indicative of the adaptability of ITL methodologies. The current ITL methodologies can be vaguely divided into two types based on how they apply the weighing strategy to the source domain instances. The first one involves re-weighing all the source instances at once using techniques such as Kernel Mean Matching (KMM)~\cite{huang2006correcting, cortes2014domain}, Weighted-Kernel Ridge Regression~\cite{garcke2014importance}, Kullback-Leibler Importance Estimation~\cite{sugiyama2008direct}, translating training instances to an Invariant Hilbert Space~\cite{herath2017learning}, or learning source domain instance weights based on the conditional distribution difference from the target domain~\cite{chattopadhyay2012multisource}. The second type of methodology is the ensemble learning methodology, primarily including the boosting techniques.

\subsection{Boosting}
Boosting~\cite{freund1999short} is an ensemble technique that builds a classifier by using a set of weak learners, whereby the weights of the training samples are updated over a chosen number of iterations, and finally these weak learners are combined to generate a strong learner. Popular boosting methodologies such as \adaR{}~\cite{drucker1997improving} typically assume that the test and training datasets have a similar distribution and hence do not require domain adaptation. They do not suffer from overfitting~\cite{sun2006reducing}, and have a robust prediction over diverse datasets.

\subsubsection{Boosting for Transfer Learning}
\trada{}~\cite{dai2007boosting} is a classification boosting framework that applies transfer learning to compensate for a lack of training instances for the target dataset. The source and target data instances are merged to form the training data for the~\trada{}, and in each iteration the weights of the instances are adjusted such that the misclassified target instances have their weights increased, whereas the misclassified source instances have their weights reduced, in order to reduce their impact towards the model learning. However, this may lead to model over-fitting, reduction in variance of the training model, therefore negatively affecting the model generalizability~\cite{wang2019transfer}.

\subsubsection{Boosting for Regression Transfer}
\tradaR{}~\cite{pardoe2010boosting} builds upon \trada{}~\cite{drucker1997improving} for regression problems, using adjusted error over residuals and reweighing of the instances. The improved version, called two-stage TrAdaBoost.R2 (\ttradaR{}), is divided into two stages. The first stage involves gradually reducing the weights of the source instances until a certain cross-validation threshold is achieved. In the second stage, weights of the source instances are frozen while the weights of the target instances are updated as in~\adaR{}. The bi-update methodology for~\ttradaR{} helps reduce the skewness produced due to source instances. This mostly happens in the cases when source sample size is very large compared to the target sample size, which consequently makes the model learning biased towards the source domain.

\subsection{Variants of Regression Transfer}
Pardoe~\etal~\cite{pardoe2010boosting} introduced two categories of transfer learning algorithms. The first category contains algorithms that choose the best hypothesis from a set of experts, each representing the models for the corresponding source dataset. This category includes algorithms such as \emph{ExpBoost.R2} and \emph{Transfer Stacking}. Algorithms in the second category, which include~\tradaR{} and~\ttradaR{}, use the grouped source and target datasets to perform boosting. Since boosting methodologies involve instance reweighing, they fall under the category of transfer learning algorithms that use domain adaptation. This is especially useful and applicable for real-world datasets with dissimilar domain distributions. Hence, such domain adaptation transfer methodologies help in reducing the burden of maintaining expert systems~\cite{rosenstein2005transfer}. Apart from the boosting methodologies, the varying domain adaptation approaches include using a kernel-employing Gaussian process~\cite{cao2010adaptive} for source instance modification or kernel ridge regression, and discrepancy minimization for domain adaptation~\cite{cortes2014domain}. Similar to importance sampling~\cite{pan2009survey}, several studies~\cite{ngiam2018domain,garcke2014importance} have used importance weighting of source instances to improve inference for transferring knowledge. Transfer methodologies using approaches similar to active learning, such as~\cite{davis2009deep} (employing modeling structure with second-order Markov chains), as well as the burgeoning variety of deep learning approaches~\cite{bengio2012deep,dauphin2012unsupervised}, are indicative of the usefulness of active learning in the form of importance sampling as a viable technique to be picked up by ITL methodologies.

\subsection{Importance Sampling}
Importance sampling is a methodology based on the concept that certain instances of the source dataset are more similarly distributed to the instances in the target dataset and thus should be sampled for learning optimal transfer models. The core tenet of importance sampling is that models should be trained with some cognizance of a multi-domain transfer, in order to avoid stale training data~\cite{katharopoulos2018not,pan2009survey,xu2019unsupervised}. Zhao \etal~\cite{zhao2015stochastic} introduce stochastic optimization for importance sampling of non-transfer learning problems, to reduce variance and improve convergence. Schuster \etal~\cite{schuster2015gradient} uses Monte Carlo methods to introduce adaptive importance sampling methodology for transfer learning. Salaken \etal~\cite{salaken2019seeded} present a seeded sampling technique for transfer learning that we extend to form the variance sampling component used by our algorithm,~\us{}. Their work introduces an algorithm to cluster the source domain instances which are then translated to limited target domain instances for knowledge/domain adaptation. In the following section, we describe how we utilize the concept used by seeded sampling for cherry picking instances from the source domain for the purpose of introducing variance in the target dataset.

\section{Methodology}
\paragraph{Problem Definition:} Given source and target datasets, such that their instances are denoted by $x^T$ and $x^S$ respectively. Hence, the target dataset is denoted as $X^T= \{x_1^T, x_2^T, ..., x_m^T\}$ for m instances and source dataset is denoted as $X^S= \{x_1^S, x_2^S, ..., x_n^S\}$ for n instances. Similarly, the target output dataset is denoted as $Y^T= \{y_1^T, y_2^T, ..., y_m^T\}$ and the source output dataset is denoted as $Y^S= \{y_1^S, y_2^S, ..., y_n^S\}$. The target domain suffers from significant data deficiency and dissimilarity of distribution compared to the source domain. Our goal is to find a transfer learning approach that can use the source domain instances as a leverage for building the prediction model as well as avoiding negative transfer. The transfer learning algorithm should perform consistently well on varying domain distributions with differing complexities. 

\paragraph{Approach:} \us{}~is a transfer regression boosting algorithm which builds a model, $h_f:X\rightarrow Y$, such that $h_f$ is the final learnt hypothesis of the ensemble of hypotheses over the learning iterations, using the training data which is a combination of source and target datasets that share a similar feature space but have dissimilar distributions. Hence by this definition, the combined training dataset (source + target) can be denoted as $\{(x,y)\| x\in X^T\cup X^S, y\in Y^T\cup Y^S~\textrm{and}~X^T, X^S, Y^T, Y^S \in R^d\}$ where $d$ represents the feature space of the source and target domain.

\begin{algorithm}
    \caption{k-Center Sampling}
    \label{alg:c-sampling}

    \KwIn{$X^T$, $Y^T$, $X^S$, $Y^S$}
    \KwOut{The labeled data set $X^{VT}$ (size k).}
    
    Find~$X^C \subset X^S$~such that $X^C = \{x_1^C, x_2^C, ..., x_k^C\} $~has k samples, obtained using \emph{k-Means Clustering} on set $X^S$.\\
    
    Initialize $X^E = \phi$ (Empty-set)

    \For{$x^C \in X^C$}
    {
    Find $x^T$ such that~~$\forall x^T\in X^T$~~min $\{||X^C - x^T ||\}$ \\
    $X^E \cup \{x^T\}$
    }
    Repeat steps $3$ to $5$  and obtain set $X^{VT}$ closest to instances in set $X^E$.\\
    
    \KwRet{$X^{VT}$}
\end{algorithm}

\subsection{\us{}}
To improve the performance of~\ttradaR{}, we present~\us{} as shown in Algorithm~\ref{alg:stradaboost-r2}. There are two main areas where~\us{}~diverges from its predecessor,~\ttradaR{}; the first is applying importance sampling, and the second is the weight update strategy for~\us{}, which differs from the~\ttradaR{}. In the following sub-sections we elaborate upon these differences as well as determine the time complexity of~\us{}.

\subsubsection{Sampling}
In order to improve the prediction accuracy,~\us{}~initially samples the source dataset, $X^S$, to obtain optimal representative instances, i.e. similar instances to the target dataset, $X^T$. It applies a greedy approach for calculating the distance between source and the target instances. Such an importance sampling can be achieved by using any distance measure such as Euclidean distance, Manhattan distance, or others. For our experiments, we use the Euclidean distance (L2 norm). Hence, we find the set $X^{ES}$ such that, 
\[X^{ES} = \lVert\mathbf{x_i^S - \Bar{x}^T}\rVert~~~~~\forall x_i\in X^S\]

where $\Bar{x}^T$ is the mean of target instances, $\lVert\mathbf{.}\rVert$ is the Euclidean distance, and $\lvert X^{ES} \rvert = \lvert X^S \rvert$, i.e. they share the same cardinality. We select the top $p$ instances from $X^{ES}$ for the source dataset, which reduces the source dataset size to $X^K= \{x_1^K, x_2^K, ..., x_p^K\}$ such that $p << n$ and discard the remaining ($n - p$) instances, since they failed the similarity testing.

Furthermore, to improve the generalizability of the prediction model, we also induce variance in the target dataset whereby source instances most similar to the target instances are added using the k-Center Sampling, an approach presented in Algorithm~\ref{alg:c-sampling}. Including the most similarly distributed source samples in the target dataset improves the fit for the regressor since~\us{} focuses more on target instances than the source instances. These similarly distributed source samples behave as noise for the target distribution and thereby improve the generalization error. Even though~\ttradaR{} tries to mitigate this using its two-stage source instance penalizing process, we found that reducing the source sample size using importance sampling, as well as performing variance sampling, allows~\us{} to perform better compared to its predecessor. 

\paragraph{k-Center Sampling} is an unsupervised approach that returns $k$ centroids, where $k$ is equal to the number of source instances in the set, $X^S$ (Alg.~\ref{alg:c-sampling}). We employ k-Center Sampling in our methodology to introduce noise in the target dataset, in order to increase its variability. After the selection of centroids, the target instances closest to these centroids are selected as the representative target set, $X^C$. The source instances most similar to the representative target set are chosen as the final subset, $X^{VT}$, for inclusion into the target dataset. The k-Center Sampling methodology is presented in Alg.~\ref{alg:c-sampling}. The final size of target dataset is, $q = n + k$. For the k-Center Sampling, the time complexity is $O(N^2)$ as result of using the k-means clustering for calculating the closeness. Hence, the sampling pipeline produces new source dataset (due to Importance Sampling) and new target dataset (due to Variance Sampling) as $X^{ES}$ and $X^{VT}$ respectively.

\begin{algorithm}
\KwIn{{The labeled data sets, $X^S$ (size n) and $X^T$ (size m)}\newline
        {The number of estimators, $N$}\newline
        {The number of cross validation folds, $F$}\newline
        {Number of Steps/Iterations, $S$}\newline
        {The base learning algorithm, $learner$}\newline
        {Learning rate, $\alpha$}\newline}
        \KwOut{Final hypothesis, $h_f$}

       \SetKwProg{KwImpSamp}{Importance Sampling}{}{}
        \KwImpSamp{\newline 
           { Get $X^{ES}$ dataset containing $p$ instances most similar to $X^S$.}
        }
        
        \SetKwProg{KwVarSamp}{Variance Sampling}{}{}
        \KwVarSamp{\newline 
            {Get  $X^{VT}$ dataset (size $q$), obtained using k-Center Sampling on set $X^T$.}
        }
        
        \SetKwProg{KwInit}{Initialize}{}{}
        \KwInit{\newline
            {Initial weight vector $w^1$ as, $w_i^1= 1/(p + q)\quad for\quad 1 \leq i \leq p + q$}
        }
        
        \For{$t \gets 1$ to $S$}
        {
            Call AdaBoost.R2 with $N$ estimators and the $learner$ to obtain hypothesis $h_{t}$.
        
            Calculate the adjusted error using the hypothesis $h_{t}$ over $F$ folds as, 
            \begin{align*}
                e_i & = \abs{y(x_i) - h(x_i)} / J~~~~~\textrm{where}~J~\textrm{is,}\\
                J & = \max_{i=1}^{(p + q)}\abs{e_i}   
            \end{align*}

            Set $\Bar{\beta_t} = \eta_t/1-\eta_t$ where 
            $\eta_t = \sum_{i=1}^{p + q}w_i^t e_i^t$ and \newline 
            $\beta_t = \dfrac{q}{(p+q)} + \dfrac{t}{(S-1)}(1-\dfrac{q}{(p+q)})$.
        
            Update the weights as: 
            \[w_i^{t+1}=
            \begin{cases}
            
            \frac{w_i^t\Bar{\beta_t}^{e_i^t}\alpha}{Z_t},~~~1\leq i\leq p \\
            
            \frac{w_i^t\beta_t^{1-e_i^t}\alpha}{Z_t},~~~p\leq i\leq (p+q) \\
            \end{cases}
            \]
            
            where $Z_t =$ sum of sample weights
        
        }
        \KwRet{}{$h_f  =$ argmin\textsubscript{i} error\textsubscript{i} }
        
   \caption{~\us{}}
   \label{alg:stradaboost-r2}
\end{algorithm}

\subsubsection{Weight Update Strategy} 
We present~\us{} in Algorithm~\ref{alg:stradaboost-r2}, where we hypothesize that by updating the target weights more aggressively, the prediction model is able to mitigate the source distribution bias. This is especially useful for dissimilar source and target domain distributions, as well as when $\lvert X^S\rvert >> \lvert X^T \rvert$. We also note that~\us{} does not employ~\adaR{}'~\cite{pardoe2010boosting}, a modified version of~\adaR{} where the weights of source instances are frozen and the weights of target instances are updated based on the reweighing approach used by~\adaR{}. However, applying highly focused domain adaptation by freezing weights of source instances can greatly reduce the generalizability of the model. Hence, in~\us{}, the hypothesis is obtained by using the~\adaR{} methodology initially. The weights for the instances are then updated iteratively using the following weight equation, 

\[w_i^{t+1}= 
    \begin{cases}
    \frac{w_i^t\Bar{\beta_t}^{e_i^t}\alpha}{Z_t}, \quad 1\leq i\leq p \\
    
    \frac{w_i^t\beta_t^{1-e_i^t}\alpha}{Z_t}, \quad 1\leq i\leq (p+q) \\
    \end{cases}
\] 

In the above equation, $\Bar{\beta_t} = \eta_t/1-\eta_t$ such that $\eta_t = \sum_{k=1}^{(p+q)}w_i^t e_i^t$, and  $Z_t = \alpha/\sum_{k=1}^{(q)}w_i $
$\beta_t$, which is equivalent to the sum of resulting weight of target instances. For the above weighing strategy, the adjusted error, $e_i$ for each instance in the training dataset, allows for the penalization of the weights of instances. For a large adjusted error the source instances are punished more than the target instances, making their weights smaller. In such a case, the algorithm focuses more on target instances. As the number of iterations increase, $\beta_t$ for target instances increases, whereas for the source instances $\Bar{\beta_t}$ depends on the adjusted error. Hence, the source instances, although much more penalized than target instances, still are not aggressively fined as is done in~\ttradaR{}, which may lead to overfitting on the dataset. We can summarize this by saying that instead of allowing the gradual weight increase of the target instances, we allow them to increase without any constraint, in order to balance the skewness in weighing caused by a large number of source instances.

\subsubsection{Time Complexity for~\us{}}
The time complexity of the~\us{}~can be divided into \emph{four} parts: 
\begin{enumerate}
    \item Time complexity of importance sampling ($O_1$)
    \item Time complexity of producing a weak hypothesis ($O_2$)
    \item Time complexity of computing the error rate in~\us{} ($O_3$)
    \item Time complexity of the second-stage of~\us{} ($O_4$)
\end{enumerate}

For S iterations, time complexity can be defined as $O(S*(O_2 + O_3 + O_4))$. For our experiments, we chose decision tree as the base learner. The time complexity for creating a decision tree is $O(d*N^2*logN)$ ($O_2$), where $d$ is the dimension of the dataset, $N$ is the number of samples, and each decision is taken in $O(logN)$ time. The time complexity of computing adjusted error combined with the weight update process ($O_3$), does not increase more than $O(N)$. Finally the time complexity of the computing the second-stage of the~\us{} is similar to producing a weak hypothesis ($O_4$). Hence, the time complexity over $S$ iterations is,

\begin{equation*}
    \begin{split}
        O(S*(d*N^2*logN + N + d*N^2*logN)) = \\
        O(2*S*d*N^2*logN + S*N)) \\
                            = O(S*d*N^2*logN))
    \end{split}
\end{equation*}

For the k-Center Sampling, the time complexity is $O(N^2)$ for calculating closeness using the k-means clustering, as well as using Manhattan distance for finding the most similar source instances. Hence, the total time complexity for ~\us{}~can be calculated as,

\begin{equation*}
    \begin{split}
        O(S*d*N^2*logN + N^2) & = O(S*d*N^2*logN) \\
    \end{split}
\end{equation*}

\subsection{Complexity of Distribution}
\label{sec:datasetcomplexity}
Domain-agnostic characterizations of dataset complexity are surprisingly uncommon. Fernandez \etal~\cite{fernandez2014information} present a characterization based on Shannon entropy, but this does not extend to the continuous, often real-valued domains of many real-world datasets~\cite{branchaud2019spectral}. Other intuitive measures such as sorting datasets by number of features or self-similarity do not reliably capture types of datasets that we observed as being especially prone to negative transfer. Heterogeneity and complexity of datasets usually determine the model performance. While heterogeneity of real-world datasets can be outlined as a factor of their multi-source and spatio-temporal character, this might not be true for their complexity. Ho \etal~\cite{ho2002complexity} proposed metrics to measure complexity for classification datasets. Maciel~\etal~\cite{maciel2016measuring} extended that work for regression datasets which later stemmed the work done by Lorena~\etal~\cite{lorena2018data} that uses meta-features as a measure of complexity. 

\begin{table*}[th]
    \centering
    \scriptsize
       %\caption{The table shows complexity of each dataset based on the $3$ complexity metrics that measure the correlation $C_{FE}$, linearity $D_{L}$ and smoothness $D_{I}$ for the distribution. For each metric, a higher value indicates a more complex distribution. From the table we observe that \emph{Kinematics} have the highest complexity ($2$ out of $3$ times) when compared to the other datasets.}
       \caption{Dataset Statistics [Tr: Training, Tt: Test, $P^M_C$: predictor] and Complexity (Section~\ref{sec:datasetcomplexity})}%moderately correlated predictor]}
    % \footnotesize
    \begin{tabular}{lrrrrrrrr} 
        %\multirow{2}{*}{\em Complexity}    & \multicolumn{8}{|c|}{UCI Dataset} \\ \cline{2-9}
 	  \toprule
       % & \multicolumn{1}{c|}{Concrete}	& \multicolumn{1}{c|}{Housing}
        %& \multicolumn{1}{c|}{Auto}       & \multicolumn{1}{c|}{Ailerons} 
        %& \multicolumn{1}{c|}{Elevators} 
        %& \multicolumn{1}{c|}{Abalone}    & \multicolumn{1}{c|}{Kinematics} 
        %& \multicolumn{1}{c|}{C.Activity} \\ \hline
 	    &Concrete&Housing&Auto&Ailerons&Elevators&Abalone&Kinematics&C.Activity\\
 	    \midrule
 	    Shape & (1030,~9)& (506,~14)& (392,~8)& \begin{tabular}{@{}c@{}}Tr: (7154,~41) \\ Tt: (6596,~41)\end{tabular}& \begin{tabular}{@{}c@{}}Tr: (8572,~19) \\ Tt: (7847,~19)\end{tabular}& (4177,~9)& (8192,~9)& (8192,~22) \\ % \hline
 	    
 	    Target & Strength& medv& mpg& goal& Goal& Rings& y& usr \\ % \hline
 	    
 	    $P^M_C$ & Cement& nox& h.power& \textit{None}& \textit{None}& weight& theta7& pgin \\ % \hline
 	    
 	    \midrule
        $C_{FE}$ & 0.66& 0.39& 0.51& 0.47& 0.59& 0.69& \textbf{0.70}& 0.36 \\ % \hline
        
        $D_{L}$  & 0.20& 0.29& 0.24& 0.26& 0.32& 0.27& 0.19& \textbf{0.36} \\   %\hline

        $D_{I}$ & 0.71& 0.90& 0.58& 0.68& 0.59& 0.51& \textbf{1.08}& 0.58    \\
         %\hline	  
        \bottomrule
 	
    \end{tabular}

    \label{tab:complexity-stdreg}
\end{table*}

\subsubsection{Collective Feature Efficiency ($C_{FE}$): Correlation Measure}
Correlation Measure determines the highly correlated predictor to the target variable and fits a linear regressor to find its residuals. All the instances having residual less than a certain threshold ($\epsilon\leq 0.1$) are discarded and the remaining instances are used to determine the next highly correlated predictor. The process is repeated until the complete feature space has been visited. Maciel et al.~\cite{maciel2016measuring} describes the measure in detail where the Collective Feature Efficiency ($C_{FE}$) is quantified as, 

\[ C_{FE} = 1 - \sum_{k} \frac{N_k}{N} \]

where $N_k$ is the number of instances that are removed (using the set threshold), $N$ is the total number of instances and $k$ is the feature. Higher values for $C_{FE}$ indicate more complex problems. 

\subsubsection{Distance from Linear Function ($D_{L}$): Linearity Measure}
Linearity Measure sums the absolute values of residuals when a multiple linear regressor is used as the learner~\cite{maciel2016measuring}. It is expressed as a distance measure ($D_{L}$) and is quantified as,

\[ D_{L} = 1 - \sum_{i=1}^{N} \frac{R_i}{N} \]

where $R_i$ are the residues and $N$ is the sample size. Lower values indicate a simpler distribution.

\subsubsection{Input Distribution ($D_{I}$): Smoothness Measure}
Smoothness Measure determines the smoothness of the distribution by ordering the predictor values in ascending order with regard to the output variable. It then finds the distance (L2 Norm) between each pair of instances~\cite{maciel2016measuring}. Lower values mean a simpler distribution, indicating that the instances in input space are closer to each other, leading to a smooth distribution. It is expressed as,

\[ D_{I} = \frac{1}{N}\sum_{i=2}^{N} \lVert\mathbf{x_i - x_{i-1}}\rVert\]

where $N$ is the sample size and $\lVert\mathbf{.}\rVert$ is the Euclidean distance.

\section{Evaluation}
For our experiments, we evaluate~\us{}~against other competitive transfer learning methodologies --~\ttradaR{}~\cite{pardoe2010boosting},~\kmmx{}~\cite{huang2006correcting},~\kliepx{}~\cite{sugiyama2008direct} and~\iwkrrx{}~\cite{garcke2014importance}  known to perform well for regression based instance transfer learning problems. Since~\ttradaR{} is the predecessor for~\us{}, we define it as the baseline algorithm for comparison. The decision tree regressor was chosen as the base learner for these methodologies. For~\ttradaR{}~and~\us{}, the following values were considered: $S$ (no. of steps) = $30$, $F$ (CV-folds) = $10$, learning rate = $0.1$ and a \textit{squared loss}. Similar values were used by Pardoe et al.~\cite{pardoe2010boosting} for their study on regression boosting. For the remaining algorithms, we used the default values for the parameters. The values were chosen to maintain generalizability of the predictions across the algorithm. They were derived using multiple experiments and iterations involving parameter tuning, and were judged to not be biased towards a single model to the best of our knowledge. The results along with the ablation study are presented in the following sections.

% \begin{table*}[t]
%     \centering
%     \scriptsize
%     % \footnotesize
%     \caption{The table shows the statistics such as- dataset shape (\emph{Shape}), output variable (\emph{Target}) and the moderately correlated predictor ($P^M_C$) for the UCI regression datasets.}
%     \begin{tabular}[t]{|c|ccc|} \hline
%         \multirow{2}{*}{\em Datasets}    & \multicolumn{3}{|c|}{\textbf{Statistics}} \\ \cline{2-4}
 	  
%         & \multicolumn{1}{|c}{\textbf{Shape}}	& \multicolumn{1}{c}{\textbf{Target}}
%         & \multicolumn{1}{c|}{$P^M_C$}  \\ \hline
 	
%         \textbf{Concrete} & (1030,~9)& CompressiveStrength& Cement  \\ \hline
   
%         \textbf{Housing}  & (506,~14)& medv& nox   \\ \hline

%         \textbf{Automobile} & (392,~8)& mpg& horsepower    \\ \hline	  
        
%         \textbf{Ailerons} & \begin{tabular}{@{}c@{}}Train: (7154,~41) \\ Test: (6596,~41)\end{tabular}& goal& \textit{None}     \\ \hline	  
        
%         \textbf{Elevators} & \begin{tabular}{@{}c@{}}Train: (8572,~19) \\ Test: (7847,~19)\end{tabular}& Goal& \textit{None}    \\ \hline	  
        
%         \textbf{Abalone} & (4177,~9)& Rings& Whole\_weight    \\ \hline	  
        
%         \textbf{Kinematics} & (8192,~9)& y& theta7    \\ \hline	  
        
%         \textbf{C.Activity} & (8192,~22)& usr& pgin    \\ \hline	  
        
%     \end{tabular}
%     \label{tab:stdreg-stats}
% \end{table*}

\paragraph{Datasets}
We chose $8$ standard regression datasets from the UCI machine learning repository~\cite{asuncion2007uci} as shown in Table~\ref{tab:complexity-stdreg}. UCI datasets were divided into \emph{source}, \emph{target}, and \emph{test} sets using the splitting methodology used by Pardoe \etal~\cite{pardoe2010boosting}. The splits were made by identifying the feature moderately correlated with the target variable, which allowed for concepts to be significantly different from each other. The \emph{first} split was considered as the target dataset and the remaining splits as the source datset. This was done so that the source sample size would be higher than the target sample size. The target dataset was further split into training and testing datasets using a K-Fold split over $20$ iterations. Our initial study showed that the Root Mean Square loss on \emph{Concrete}, \emph{Housing}, and \emph{Automobile} datasets were moderately varied for such a division which allowed for robust predictions since it incorporated both generalizability for the models, as well as lesser noise. Hence, we further extended the splitting methodology to other datasets- \emph{Abalone}, \emph{Kinematics} and \emph{Computer Activity}. For \emph{Ailerons} and \emph{Elevators} datasets, the UCI repository already consisted of a testing dataset. We took very few target instances so that the remaining larger dataset could be used as the source dataset, which in turn imitates a real world transfer learning problem. Table~\ref{tab:complexity-stdreg} shows the dataset statistics including their size, target variable, and predictor used for correlation splitting. 
Although Concrete, Housing, and Automobile are small sample datasets, they were used to imitate the study by Pardoe et al.~\cite{pardoe2010boosting}. We compensated for this imbalance using other large sample datasets with varying heterogeneity. The complexity evaluation in Table~\ref{tab:complexity-stdreg} shows how complex ($C_{FE}$ \ie variance, $D_{I}$ \ie smoothness, and $D_{L}$ \ie linearity) these distributions are. For each metric, a higher value indicates a more complex distribution. From the table we observe that \emph{Kinematics} has the highest complexity ($2$ out of $3$ times) when compared to the other datasets.

\paragraph{Ablation Study}
We perform the Ablation study where importance sampling technique is applied individually to each transfer learning methodologies. The goal of this study is to induce fairness in comparison, given the modular nature of importance sampling technique. Sampling is a two-phase methodology which includes Variance Sampling and importance sampling. Variance Sampling includes sprinkling the target dataset with source instances in order to introduce noise and increase the variance of the distribution. For the Concrete, Housing and Automobile datasets, Variance Sampling was not applied due to low sample size. Importance sampling on the other hand uses similarity measuring to find the source instances most similar (important) to the target instances. The Ablation study exploits importance sampling for all the algorithms and Variance sampling for datsets (not including the small datasets).

\begin{figure*}[hbtp]
    \centering
    \includegraphics[width=\textwidth]{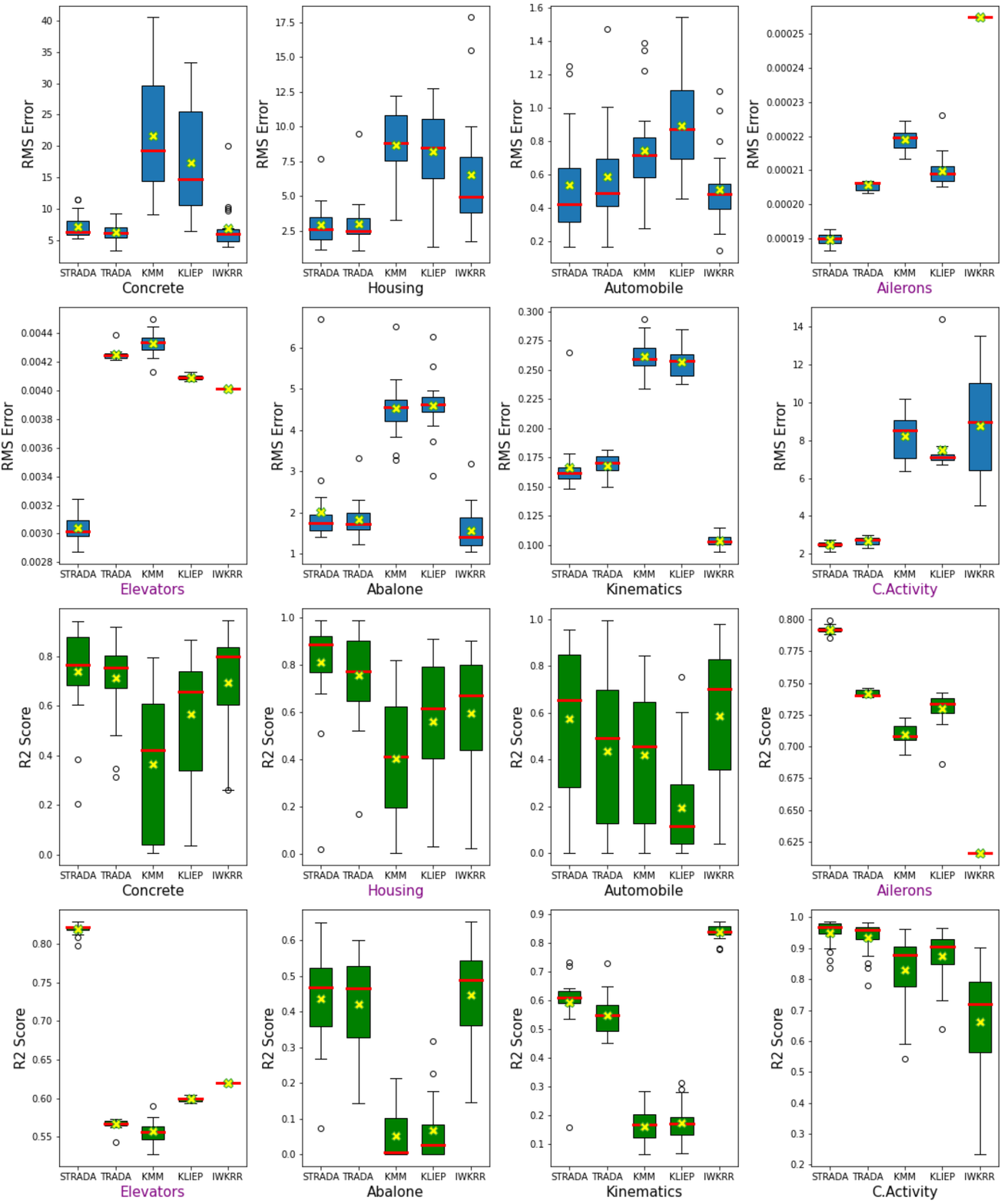}
    % {paper_r2_rmse_stdreg.eps}
    %{paper_r2_rmse_stdreg.png}
    \caption{Comparison of transfer learning algorithms-- TRADA:~\ttradaR{}, STRADA:~\us{}, KMM:~\kmmx{}, and KLIEP:~\kliepx{}, IWKRR:~\iwkrrx{}, where the RMS error and R-squared score is calculated over 20 iterations. The Interquartile Range (IQR), mean value (marker: yellow "X"), and median value (marker: red line) for each algorithm over the iterations have been highlighted. The datasets for which~\us{} performs particularly well are marked as well (marker: purple).}
    %Marker Legend for mean and median.
    \label{fig:uci-rmse-r2}
\end{figure*}

\subsection{Results}
%Importance Sampling study for 3 datasets [C.act, kinematics, ablaone (1/8, 1/4, 1/2, 1)]
We implemented the experiments on an HPC cluster with $16$ processors and $128$ GB RAM. Any required short supplemental processing was performed on personal laptops with half the number of processors and RAM. The number of cross-validation fold iterations were $20$ for the datasets. The distribution of prediction values is shown in the box-plot Fig.~\ref{fig:uci-rmse-r2}. We observe that~\us~consistently performs well, with low RMSE as well as high R-squared score. However, this is not true for other methodologies, especially~\iwkrrx{} and~\ttradaR{} which, although they sometimes outperform~\us{}, they also fluctuate highly in their performance. Example~\iwkrrx{}~is the most optimal model for Automobile, Abalone and Kinematics datasets as observed through its mean RMSE and R-squared values. But it is not consistent in its performance as observed for C.Activity, Ailerons and Elevators datasets, where it fluctuates highly in its mean and variance over the iterations. However,~\us{} performs consistently well for all of the datasets and comes a close second in the Kinematics dataset, where~\iwkrrx{} outperforms the competing methodologies by a high margin. Similarly, for~\ttradaR{}, we observe that it performs well (RMSE score) on Concrete and Abalone datasets compared to~\us{}, but its performance is not consistent as observed for Ailerons and Elevators datasets. We consider~\ttradaR{} to be our baseline algorithm for this study primarily because it is the predecessor of~\us{}, and observe that~\us{} outperforms~\ttradaR{} $75\%$ of the times in the case of loss measure, and $100\%$ when measured for correlation accuracy.

\begin{table*}[ht]
    \centering
    \scriptsize
    % \footnotesize
    \caption{Ablation Study}% for the UCI regression datasets. We compare the transfer learning algorithms where Importance Sampling was used for each of the algorithms-  TRADA:~\ttradaR{}, STRADA:~\us{}, KMM:~\kmmx{}, and KLIEP:~\kliepx{}, IWKRR:~\iwkrrx{}. The table provides the mean RMS error and R-squared score calculated over 20 iterations.}
    \begin{tabular}[t]{lrrrrrrrrrr} %\hline
       % \multirow{2}{*}{\em Algo.}    & \multicolumn{10}{|c|}{UCI dataset} \\ \cline{2-11}
 	  
        %& \multicolumn{2}{c|}{Ailerons}	& \multicolumn{2}{c|}{Elevators}
        %& \multicolumn{2}{c|}{Abalone}    & \multicolumn{2}{c|}{Kinematics} 
        %& \multicolumn{2}{c|}{C.Activity} \\ %\hline
        \toprule
 	&Ailerons&&Elevators&&Abalone&&Kinematics&&C.Activity\\
 	    \midrule
 	    & RMSE & $R^2$& RMS& $R^2$& RMS& $R^2$& RMS& $R^2$& RMS& $R^2$  \\ %\hline
 	
        TRADA &  0.00023& 0.65& 0.0042& 0.38& 2.14& 0.40& 0.18& 0.47& 2.98& 0.92  \\ %\hline

        STRADA & \textbf{0.00018}& \textbf{0.79}& 0.0030& \textbf{0.81}& 2.02& \textbf{0.43}& 0.18& 0.51& \textbf{2.48}& \textbf{0.94}  \\ %\hline

        KMM & 0.00029& 0.46& 0.0049& 0.31& 2.73& 0.06& 0.27& 0.08& 11.30& 0.17  \\ %\hline

        KLIEP & 0.00026& 0.58& 0.0043& 0.42& 2.76& 0.10& 0.26& 0.10& 11.09& 0.22  \\ %\hline
    
        IWKRR & 0.00025& 0.63& \textbf{0.0021}& \textbf{0.81}& \textbf{1.99}& 0.41& \textbf{0.10}& \textbf{0.84}& 8.77& 0.66  \\
        %\hline
        \bottomrule
    \end{tabular}
    \label{tab:abaltion-stdreg}
\end{table*}

Considering importance sampling is a pre-domain adaptation methodology and should not be limited to just~\us{}, we conduct an Ablation study as shown in Table.~\ref{tab:abaltion-stdreg}. We observe minimal improvement in the performance of~\ttradaR{}~and~\iwkrrx{} and find that~\us{}~performs consistently well (4 out of 5 times). Table.~\ref{tab:abaltion-stdreg} shows that~\iwkrrx{}~has competitive scores with regard to ~\us{}, however it suffers from the same inconsistency as observed in the comparative study presented in Fig.~\ref{fig:uci-rmse-r2}. Also,~\ttradaR{} does not show any improvement except for a similar RMSE score to~\us{} for the Kinematics dataset. However,~\iwkrrx{} easily outperforms all other methodologies for the Kinematics dataset. It should also be noted that in both the studies, the remaining algorithms-~\kmmx{} and~\kliepx{} performed quite poorly compared to the other methodologies and  showed less signs of improvement in either cases. Hence, we can say that~\us{}~has shown itself to be consistent among all the measures, adapting more robustly to more complex and varying distribution datasets. It should also be noted that~\us{} has consistently empirically outperformed, as well as being shown effective in avoiding negative transfer.~\us{}~employs~\adaR{}~as an initial step and penalizes both source and target instances while re-weighing, instead of only source instances as done in~\ttradaR{}. Consequently, this generalizes the training dataset which in turn produces consistent performance on a diverse group of datasets, as seen in prior~\cite{wang2019transfer} as well as our transfer learning research. 

\section{Discussion}
Since~\us{} is a successor to~\ttradaR{}, we take~\ttradaR{} as the baseline and observe that~\us{}~outperforms it $7$ out of $8$ times during the comparative study. We also note that~\ttradaR{} shows no significant improvement during the ablation study. This justifies the steady performance of~\us{}, where it consistently scores optimal RMSE and R-squared values during the comparative and ablation studies. The ablation study is used to justify how importance sampling is useful when combined with the learning methodology for~\us{}. This is due to balanced weighing and generalization, complimented by source domain sampling methodology. We find that for relatively complex datasets (\emph{Concrete}, \emph{Elevators}, \emph{Kinematics}, and \emph{C.Activity}) derived from the complexity analysis presented in Table~\ref{tab:complexity-stdreg} ,~\us{}~performs well on most of them ($3$ out of $4$ times), falling short only in the case of Kinematics dataset when compared to~\iwkrrx{} methodology. 

It should be noted that both the training error and the generalization error of a similar problem space are analyzed thoroughly in Freund~\etal~\cite{freund1999short}, and this analysis is further known to apply to \tradaR~\cite{pardoe2010boosting}, a predecessor to~\us{}. The objective function for transfer learning involves minimizing the loss, $min_h\{\mathcal{L}(h) + \lambda\eta\}$, where $\eta$ is the regularization function, and $\lambda$ is the regularization constant for the loss function $\mathcal{L}$. We hypothesize a function $h \in H$ that maps training instances, predictor $x\in X$ to target $y\in Y$ in the target domain $T_T$.
Hence, the instance transfer methodology tries to minimize the weighted loss of target and source domain~\cite{wang2019transfer}, such that, $\mathcal{L}(h) = \mathcal{L}_T(h) + \mathcal{L}_S(h)$. Since,~\us{}~relies on using~\adaR{}~unlike~\ttradaR{}~\cite{pardoe2010boosting}, it has increased generalizability as it avoids overfitting while assigning balanced source and target weights.

\section{Conclusion}

We introduce~\us{}, which uses importance sampling combined with an unrestricted weight update strategy to improve performance for the domain of instance transfer learning by an average of $12 \%$ across all datasets, and $13 \%$ in sufficiently complex datasets when compared to its predecessor~\ttradaR{}. To better characterize the datasets that~\us{}~performs well on, we provide complexity measures, $C_{FE}$, $D_{L}$ and $D_{I}$ that employ feature correlation and fitting a linear regressor for computation of complexity for distributions. Hence, we can conclude that~\us{}~would be well suited for complex real-world datasets that range in distributions, as well as uniformity of features.  While the functional improvement is large, the additional overhead and physical changes we propose to~\ttradaR{} are modest enough that we expect~\us{}~to function as a drop in replacement for~\ttradaR{} and other instance transfer methodologies in scientific data analysis pipelines. Wide adoption of~\us{}~will allow for accurate inference in areas previously deemed too data-poor for modeling and too dissimilar for transfer, opening the door for new insights across complex scientific domains. We intend to release our Python and R implementations of~\us{}~on GitHub under a Creative Commons license upon publication.

\section{Future Work}
In the future, we want to expand our methodology to not employ not just instance based transfer learning methodologies but also feature based~\cite{NIPS2006_0afa92fc, argyriou2007spectral, jebara2004multi} as well as parameter based techniques~\cite{lawrence2004learning, schwaighofer2004learning}. Although boosting transfer methodologies are simpler to understand than their deep learning counterparts, the user may suffer a trade off in  prediction accuracy for simplicity, which is not always preferred. We also plan to compare boosting based instance transfer learning methodologies to deep transfer learning methodologies~\cite{zhang2018importance}. We plan to explore a methodology that uses performance gap minimization to improve the boosting in transfer learning, extending on the work of~\cite{wang2019transfer}. Complexity of distribution also plays an important part in providing a glimpse of how distributions, as well as predictions, vary. Hence, we plan to investigate other implications of characterizing data by cross-feature complexity, particularly techniques involving correlation to optimal tree depth for network learning models of data.

%Complete abstract.
%Add references in Future work.
%Add 4-5 lines in the introduction.
%Go over discussion and Conclusion.

\bibliographystyle{bst/sn-basic}
\bibliography{sn-bibliography}% common bib file


\end{document}